\pdfoutput=1
\documentclass[10pt,twocolumn,letterpaper]{article}

\usepackage{cvpr}          % camera-ready: no [review]
\usepackage{times}
\usepackage{epsfig}
\usepackage{graphicx}
\usepackage{amsmath}
\usepackage{amssymb}
\usepackage{booktabs}
\usepackage{multirow}
\usepackage{placeins}
\usepackage[pagebackref=true,
            colorlinks=true,
            linkcolor=blue,
            citecolor=blue,
            urlcolor=blue,
            bookmarks=true]{hyperref}
\usepackage{cleveref}

\begin{document}

%%%%%%%%% TITLE
\title{Long-Range Video Presentation Attack Detection: A Three-Class
Benchmark Across 1--100\,m and Two Sensor Types}

\author{Lambert Igene$^{1\dagger}$\quad
Mohammad Zahir Uddin Chowdhury$^{2}$\quad
Afzal Hossain$^{2}$\quad\\
Sandip Purnapatra$^{2}$\quad
Peter Johnson$^{2}$\quad
Stephanie Schuckers$^{1}$\\[4pt]
$^{1}$Clarkson University, NY, USA\quad
$^{2}$Ombra LLC, NY, USA\quad
$^{1}$UNC at Charlotte, NC, USA\\
{\tt\small $^{1\dagger}$igenela@clarkson.edu},
\textsuperscript{2}{\tt\small mochowd@clarkson.edu},
\textsuperscript{2}{\tt\small afhossa@clarkson.edu},
\textsuperscript{2}{\tt\small purnaps@clarkson.edu},\\
\textsuperscript{2}{\tt\small peter@ombra.us},
\textsuperscript{1}{\tt\small sschucke@charlotte.edu}
}

\maketitle
\thispagestyle{empty}

%-----------------------------------------------------------------------
\begin{abstract}
Face presentation attack detection~(PAD) systems deployed in border
control and airport surveillance are constrained by a critical
assumption: subjects are captured at close range, typically under one
meter. No existing benchmark systematically addresses multi-distance
capture, diverse environmental conditions, and concurrent spoofing
and occlusion-based attacks, leaving deployed models underprepared
for real-world scenarios. We introduce a video-based PAD dataset of
374,000+ frames from 261 unique identities collected across five
standoff distances~(1\,m--100\,m) under indoor and outdoor conditions
using a DSLR camera~(C) and a mobile sensor~(M), with three-class
labels: live, spoof, and props~(occlusions); models are trained and
evaluated on video clips rather than single frames. We fine-tune and
benchmark four CNN architectures: MobileNetV3, MobileNet, EfficientNet,
and ResNet-50. The higher-resolution DSLR maintains near-perfect spoof
classification rate across all distances, while the lower-resolution
mobile sensor degrades sharply beyond 50\,m, with props classification
rate dropping to 55\% at 75\,m for MobileNetV3. EfficientNet achieves
the best overall performance~(98.48\% accuracy, ACER 2.01\%).
Generalization tests on unseen attack datasets yield 89--94\% accuracy,
validating cross-domain robustness. These results establish capture
distance and sensor resolution as critical but underexplored axes of
PAD difficulty. The dataset will be publicly released under an
End-User License Agreement to support reproducible long-range PAD
research.
\end{abstract}

%-----------------------------------------------------------------------
\section{Introduction}
\label{sec:intro}

According to the 2024 IATA Global Passenger Survey, 46\% of airport
passengers used biometrics in 2024~\cite{regulaforensics2025biometrics},
reflecting the rapid integration of face recognition into border-control
workflows. The security of such deployments rests on the ability of
presentation attack detection~(PAD) models to reject printed
photographs, video replays, 3D masks, and occlusion-based evasion to
prevent impostors from passing as enrolled identities. Yet virtually
all PAD algorithms are trained and evaluated at close range~($<$1\,m)
under controlled conditions. Operational deployments may need to detect
spoofing and occlusion-based evasion across standoff distances from one
to over one hundred metres, under varying indoor and outdoor lighting.
A 2024 U.S.\ DHS report acknowledged that no viable camera solution
yet reliably captures biometrics at distance in real
time~\cite{dhsoig2024}.

Existing benchmarks like NUAA~\cite{wen2015nuaa},
Replay-Attack~\cite{chingovska2012lbp},
CASIA-FASD~\cite{zhang2019casia}, OULU-NPU~\cite{boulkenafet2017oulunpu}, and Livdet Face datasets ~\cite{igene2024development,igene2024face} were all captured within 0.5--1.5\,m.
As distance increases, face resolution degrades and the fine-grained
texture cues on which CNNs rely become
indistinguishable~\cite{sciopen2025cnn}. The CVPR~2023 FAS Workshop
identified long-distance PAD as insufficiently
explored~\cite{cvpr2023fascvpr}, and occlusion compounds the problem:
props such as hats and masks not only obscure liveness cues but can
constitute attack instruments, yet their interaction with distance
remains unstudied.

Critically, no prior work has studied the joint effect of occlusion
and capture distance on PAD performance, a combination that is
operationally inevitable in border control and airport surveillance
settings where a subject wearing a hat or mask may be detected at
50\,m or beyond. Our three-class formulation, which explicitly
distinguishes live subjects, spoofs, and props~(occlusion items),
enables this analysis for the first time. We address this gap with
a video-based PAD benchmark that jointly varies three factors prior
work has not controlled together: \emph{capture distance},
\emph{sensor resolution}, and \emph{environmental condition}. The
dataset spans five distances~(1\,m, 10\,m, 50\,m, 75\,m, 100\,m),
indoor/outdoor settings, and two sensor types, yielding 374,000+
labeled frames from 250 identities across three classes: live, spoof,
and props. Our contributions are:
\begin{itemize}\setlength\itemsep{0pt}
  \item A multi-distance, multi-condition video PAD dataset with
        three-class labels across two sensor types, to be publicly
        released under a EULA.
  \item The first systematic study of capture distance and occlusion
        as joint axes of PAD difficulty.
  \item A CNN benchmark (MobileNetV3, MobileNet, EfficientNet,
        ResNet-50) with class-wise recall across all
        distance--condition--sensor combinations, plus cross-dataset
        generalisation analysis.
\end{itemize}

%----------------------------------------------------------------------
\section{Related Work}
\label{sec:related}

\subsection{Face PAD Datasets and Benchmarks}
\label{subsec:datasets}

In Table~\ref{tab:dataset_comparison} benchmarks such as NUAA~\cite{wen2015nuaa},
Replay-Attack~\cite{chingovska2012lbp}, and
CASIA-FASD~\cite{zhang2019casia}, established the binary
live-vs-spoof paradigm. MSU-MFSD~\cite{wen2015nuaa} extended
coverage to mobile capture; OULU-NPU~\cite{boulkenafet2017oulunpu}
introduced lighting and device variation. A consistent limitation is
their close-range acquisition protocol: subjects sit within 0.5--1.5\,m
of the sensor. More recent work has broadened attack diversity: the
Wild Face Anti-Spoofing dataset~(WFAS)~\cite{wang2023wildface} spans
17 attack types from internet-sourced data, and the LivDet-Face 2024
competition~\cite{igene2024face} advanced standardised evaluation
across diverse instruments, but still at close range. Across all
these efforts, capture distance has not been incorporated as a
controlled experimental variable, and prevailing evaluation protocols
collapse video sequences to per-frame classification, foregoing
temporal liveness cues that video capture would support.
Table~\ref{tab:dataset_comparison} summarises key datasets alongside
ours.

\subsection{CNN Architectures for PAD}
\label{subsec:cnn}

CNNs have been the dominant paradigm for face PAD since Atoum
et~al.~\cite{atoum2017patch} showed that patch-based CNN approaches
outperform handcrafted texture features. Transfer learning from
ImageNet-pretrained models is now standard~\cite{lucena2017transfer}.
Al-Smadi et~al.~\cite{sciopen2025cnn} benchmarked eight pretrained
architectures on NUAA and Replay-Attack, finding that fine-tuning
substantially improves all models. Luevano
et~al.~\cite{luevano2024mobilenetv3} showed that preprocessing and
fine-tuning strategy outweigh architectural choice for MobileNetV3.
Beyond CNN backbones, transformer-based and generative approaches
have emerged as active directions~\cite{wang2025fassurvey}. We
restrict the present benchmark to CNN backbones because of operational
border-control constraints such as, real-time inference on edge, and
commodity hardware, makes lightweight CNNs the most immediately
deployable baseline; comparison against transformer and generative
approaches under the same distance-controlled protocol is left to
future work.

\subsection{Domain Generalisation in Face PAD}
\label{subsec:domgen}

Domain shift is a central PAD challenge. Early approaches employed
adversarial learning~\cite{jia2020ssdg} and meta-learning to learn
domain-invariant representations. Zhou et~al.~\cite{zhou2023iadg}
proposed instance-aware generalisation; Liu
et~al.~\cite{liu2023unsupervised} introduced unsupervised
generalisation via self-supervised pretraining. None of these
approaches explicitly models capture distance as a domain variable.
Yet distance-induced quality degradation produces qualitatively
distinct shifts~\cite{survfas2023}: the face region shrinks, texture
cues coarsen, and background context grows in relative proportion.
Our approach addresses this through deliberate training-data
variability across all five distance levels.

\subsection{Long-Range Biometric Recognition}
\label{subsec:surveillance}

Long-range face \emph{recognition} has received attention through
the IARPA BRIAR program~\cite{cornett2023briar}, which covers
distances up to several hundred metres, and
FarSight~\cite{farsight2025}, which integrates face, gait, and body
shape for surveillance scenarios. None of these systems evaluates
liveness or spoofing resistance. Within the PAD community,
SuHiFiMask~\cite{survfas2023} is the closest prior work, defining a
binary short/long dichotomy ($<$1\,m vs.\ $>$3\,m) with no
per-distance breakdown and no props category. The CVPR~2023
Surveillance FAS Challenge~\cite{cvpr2023fascvpr} confirmed that
existing algorithms degrade significantly under
surveillance-quality imagery. Our work extends this frontier with
five controlled distance levels, paired indoor/outdoor conditions,
two sensor modalities, and an explicit props class.

\subsection{Occlusion in Face PAD}
\label{subsec:occlusion}

Occlusion is a dual threat: it degrades genuine biometric samples by
obscuring liveness cues, and it can itself constitute an attack
instrument. ISO~30107-3~\cite{iso301073} classifies high-quality
3D masks, and prosthetics as obfuscation-based PAIs. Despite this
dual role, occlusion has received limited treatment as an independent
threat category. PAD work on mask-based
attacks~\cite{george2020multichannel,etasr2025flnet} treats the mask
as a spoof rather than as a benign occlusion item worn by a live
subject. Our three-class formulation is the first to separate these
cases explicitly in a distance-controlled setting.

%-----------------------------------------------------------------------
\begin{table*}[t]
\centering
\renewcommand{\arraystretch}{1.2}
\caption{Comparison of publicly available face PAD datasets and the
proposed dataset. V\,=\,video; I\,=\,image; $\checkmark$\,=\,supported;
$\times$\,=\,not supported; ---\,=\,not reported.
``$<$1\,m'' indicates close-range capture with no controlled distance
protocol. $\dagger$\,Subject count drawn partially from web sources.}
\label{tab:dataset_comparison}
\resizebox{\textwidth}{!}{%
\begin{tabular}{lcccccccc}
\hline
\textbf{Dataset} & \textbf{Year} & \textbf{Format} & \textbf{Subjects} &
\textbf{Live / Spoof} & \textbf{Attack Types} &
\textbf{Props/Occl.} & \textbf{Dist. Range} & \textbf{Dist. Levels} \\
\hline
NUAA~\cite{wen2015nuaa}
  & 2010 & I & 15 & 1{,}743\,/\,3{,}362 & 1 Print & $\times$ & $<$1\,m & 1 \\
Replay-Attack~\cite{chingovska2012lbp}
  & 2012 & V & 50 & 200\,/\,1{,}000 & 1 Print, 2 Replay & $\times$ & $<$1\,m & 1 \\
CASIA-FASD~\cite{zhang2019casia}
  & 2012 & V & 50 & 150\,/\,450 & 1 Print, 1 Replay & $\times$ & $<$1\,m & 1 \\
MSU-MFSD~\cite{wen2015nuaa}
  & 2015 & V & 35 & 70\,/\,210 & 1 Print, 2 Replay & $\times$ & $<$1\,m & 1 \\
OULU-NPU~\cite{boulkenafet2017oulunpu}
  & 2017 & V & 55 & 1{,}980\,/\,3{,}960 & 2 Print, 2 Replay & $\times$ & $<$1\,m & 1 \\
SiW~\cite{thapa2023siw}
  & 2018 & V & 165 & 1{,}320\,/\,3{,}960 & 2 Print, 4 Replay & $\times$ & $<$1\,m & 1 \\
CASIA-SURF~\cite{zhang2019casia}
  & 2019 & V & 1{,}000 & 3{,}000\,/\,18{,}000 & 6 Print (RGB+D+IR) & $\times$ & $<$1\,m & 1 \\
SiW-M~\cite{thapa2023siw}
  & 2019 & V & 493$^{\dagger}$ & 660\,/\,968 & Print, Replay, Mask, Makeup, Partial & $\times$ & $<$1\,m & 1 \\
WFAS~\cite{wang2023wildface}
  & 2023 & I & 469k$^{\dagger}$ & 529k\,/\,853k & 17 PAs (Print, Replay, Mask, 3D) & $\times$ & Unconstr. & --- \\
SuHiFiMask~\cite{survfas2023}
  & 2022 & V & 101 & 10{,}195\,/\,10{,}195 & 2D + 3D Mask + Adversarial & $\times$ & $>$3\,m (binary) & 2 \\
LivDet-Face 2024~\cite{igene2024face}
  & 2024 & V & --- & --- & Print, Replay, 3D Mask & $\times$ & $<$1\,m & 1 \\
\hline
\textbf{Ours}
  & \textbf{2025} & \textbf{V} & \textbf{261} & \textbf{374k+ frames}
  & \textbf{Print, 3D Mask} & $\checkmark$ & \textbf{1\,m--100\,m} & \textbf{5} \\
\hline
\end{tabular}%
}
\vspace{2pt}
{\footnotesize Two sensors: DSLR~(C) and mobile~(M). Indoor and outdoor
conditions covered at all distance levels.}
\end{table*}

%-----------------------------------------------------------------------
\section{Dataset and Protocol}
\label{sec:dataset}

\subsection{Data Collection}
\label{subsec:collection}

Data was collected at Clarkson University (Jan 2025--Feb 2026) using two
sensors: a Canon EOS~90D DSLR~(sensor~C, 32.5\,MP APS-C,
3.2\,$\mu$m pixel pitch) and a Samsung Galaxy~S24 Ultra
smartphone~(sensor~M). For the DSLR, a Sigma 150--600\,mm
f/5--6.3 DG OS HSM Contemporary telephoto zoom lens was used, set
to \textbf{600\,mm focal length} for all sessions at distance $\geq$50\,m,
and 150\,mm for distance 1\,m and 10\,m sessions. The mobile sensor used
its built-in 2$\times$ optical zoom for video capture.

Five standoff distances were measured with a tape: 1\,m, 10\,m,
50\,m, 75\,m, and 100\,m. Indoor sessions covered 1--75\,m under
stable artificial lighting; outdoor sessions covered 1--100\,m with
natural lighting variation. During video capture, subjects performed
slight lateral head movements to introduce natural motion cues.
Videos were recorded at 30 frames per second~(FPS); every 15th frame
was extracted using OpenCV~\cite{opencv2015}, corresponding to a
temporal sampling interval of 0.5 seconds~(2 frames per second).

Table~\ref{tab:sensor_resolution} reports the raw face crop size and
inter-ocular distance~(IOD) in pixels for each sensor--distance
combination, computed from the thin-lens formula using a mean face
width of 150\,mm and IOD of 63\,mm. At 100\,m, the DSLR yields an
IOD of 118\,px, exceeding the $\geq$90\,px threshold recommended
by NIST\,FRVT and ISO\,39794-5 for high-quality face
images~\cite{nistfrvt2020}, and well above the 10--20\,px IOD at
which the IARPA BRIAR programme operates at comparable
distances~\cite{cornett2023briar}. With the 40\% loose-crop
expansion~(Sec.~\ref{subsec:annotation}), the DSLR input patch at
100\,m is $\approx$393$\times$524\,px, so the resize to
384$\times$384 is a downscale at every distance; no artificial
upscaling occurs.

\begin{table}[t]
\centering
\renewcommand{\arraystretch}{1.10}
\caption{Raw face crop size and inter-ocular distance~(IOD) in pixels
per sensor and distance, computed via the thin-lens formula
(face width $\approx$150\,mm, IOD $\approx$63\,mm).
DSLR: Canon EOS~90D + Sigma 600\,mm; Mobile: Samsung S24 Ultra
2$\times$ zoom. Resize factor to 384\,px after 40\% bounding-box
expansion: values $<$1 indicate downscaling, $>$1 indicate upscaling.}
\label{tab:sensor_resolution}
\resizebox{\columnwidth}{!}{%
\begin{tabular}{llccc}
\hline
\textbf{Sensor} & \textbf{Dist.} & \textbf{Raw crop (px)} &
\textbf{IOD (px)} & \textbf{Resize factor} \\
\hline
\multirow{5}{*}{DSLR (C)}
 & 1\,m   & 28090$\times$37453 & 11798 & 0.01$\times$  \\
 & 10\,m  & 2809$\times$3745   & 1180  & 0.07$\times$  \\
 & 50\,m  & 562$\times$749     & 236   & 0.37$\times$  \\
 & 75\,m  & 375$\times$499     & 157   & 0.55$\times$  \\
 & 100\,m & 281$\times$375     & 118   & 0.73$\times$  \\
\hline
\multirow{5}{*}{Mobile (M)}
 & 1\,m   & 1446$\times$1928   & 607   & 0.14$\times$  \\
 & 10\,m  & 145$\times$193     & 61    & 1.42$\times$  \\
 & 50\,m  & 29$\times$39       & 12    & 7.11$\times$  \\
 & 75\,m  & 19$\times$26       & 8     & 10.67$\times$ \\
 & 100\,m & 14$\times$19       & 6     & 14.23$\times$ \\
\hline
\end{tabular}%
}
\end{table}

\begin{figure*}[t]
  \centering
  \includegraphics[width=\textwidth]{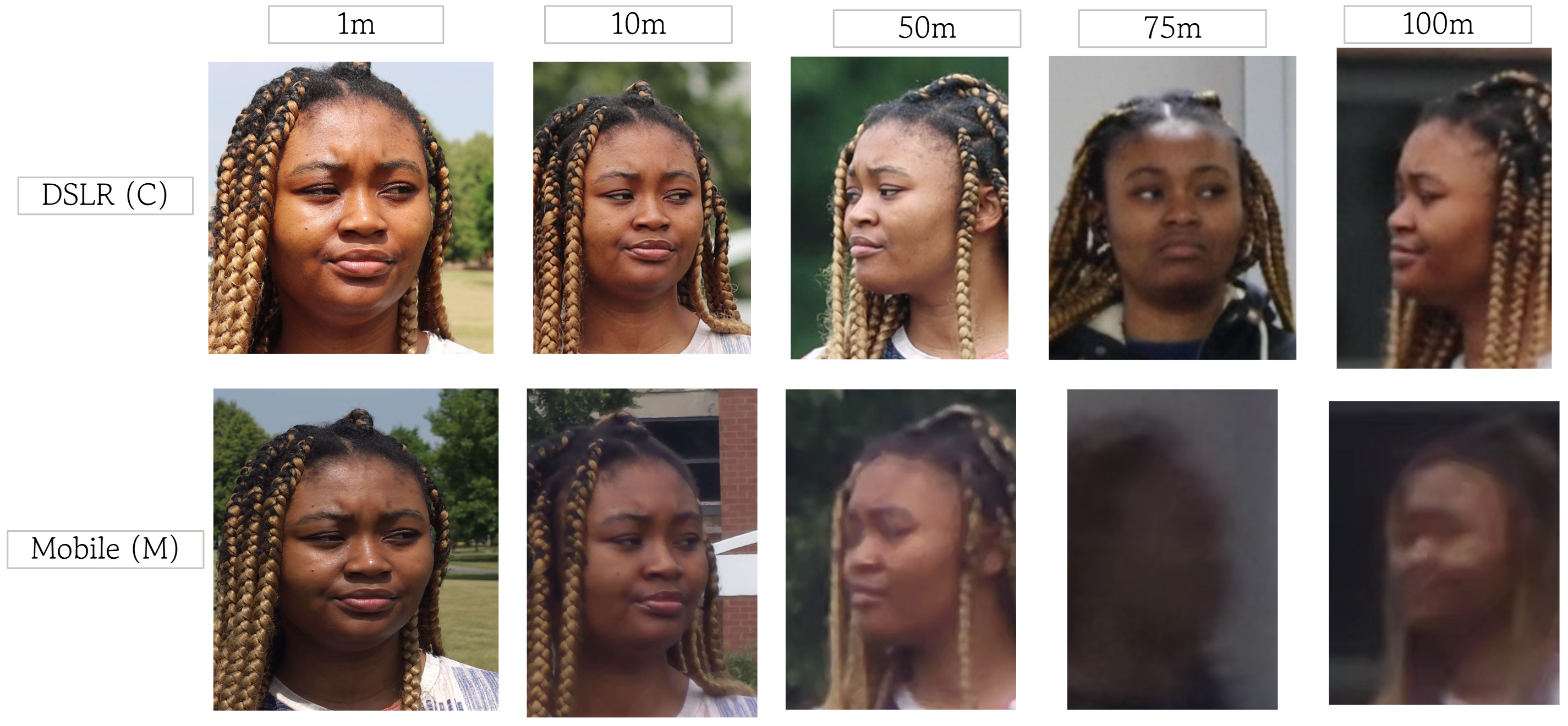}
  \caption{RetinaFace-cropped face samples for a representative subject
  across all five standoff distances, for the DSLR~(C, top row) and
  mobile~(M, bottom row) sensors. DSLR crops~(Canon EOS~90D + Sigma
  150--600\,mm at 600\,mm for $\geq$50\,m) remain sharp and
  face-centred at all distances. Mobile crops degrade progressively
  beyond 10\,m and become severely blurred by 75--100\,m, consistent
  with the IOD pixel counts in Table~\ref{tab:sensor_resolution}
  (8\,px at 75\,m and 6\,px at 100\,m for the mobile sensor).
  Bounding boxes are expanded by 40\% in both dimensions prior to
  resizing to $384\times384$~pixels for model input.}
  \label{fig:samples}
\end{figure*}

% figure files renamed without spaces for arXiv
Fig.~\ref{fig:samples} shows RetinaFace-cropped face samples across
all distances for both sensors.

\subsection{Subjects and Presentation Conditions}

The dataset includes 250 volunteer participants across 283 sessions;
33 subjects contributed both indoor and outdoor data.
\textbf{Live} --- standard captures without artifacts;
\textbf{Props} --- three controlled occlusion categories
(hair/wig, eyewear, facial coverings such as hats and scarves);
\textbf{Spoof} --- 2D attacks (printed photographs, cloth masks) and
3D attacks (rigid and silicone masks), with some spoof conditions
additionally combined with props.

\subsection{Annotation and Preprocessing}
\label{subsec:annotation}

Labels are derived directly from the acquisition protocol: every
frame from a session inherits the condition label (live, props, or
spoof). Face detection uses
RetinaFace~\cite{retinaface2019}~(InsightFace, ResNet-50 backbone),
achieving 99.6\% detection rate with zero false
positives~\cite{21362714}. A \emph{loose cropping} strategy expands each detected bounding box by
40\% in both width and height to retain surrounding props and
contextual cues critical for the three-class formulation. Where
multiple faces appear, the highest-confidence detection is retained.
Frames yielding no detection ($<$0.4\%) are excluded.

\subsection{Evaluation Protocol}

All models are trained on data from all five distances,
environments, and sensors. Performance is reported stratified by
distance, environment, and sensor to localise failure modes. The
dataset is split 80:10:10 by \emph{subject identity}~(209 train /
26 val / 26 test), ensuring no test identity appears in training. A
spoof-aware bucketing rule further prevents leakage: spoof frames are
assigned to the bucket of the \emph{mask subject}~(whose face the
mask portrays), not the wearer. Metrics: APCER, BPCER,
ACER~(ISO~30107-3~\cite{iso301073}) and per-class classification
rate.

%-----------------------------------------------------------------------
\section{Methodology}
\label{sec:method}

\subsection{Problem Formulation}
\label{subsec:formulation}

We formulate face PAD as a three-class supervised classification
problem. Given a face image crop
$\mathbf{x} \in \mathbb{R}^{H \times W \times 3}$ extracted from a
video frame, the goal is to predict
$y \in \{\text{live}, \text{spoof}, \text{props}\}$, where
\textit{live} denotes a genuine bonafide presentation, \textit{spoof}
denotes a 2D or 3D presentation attack, and \textit{props} denotes a
live subject carrying an occlusion-based prop that is not itself a
spoof instrument. This three-class formulation explicitly separates
occlusion from attack, enabling per-class error analysis and exposing
failure modes that binary PAD benchmarks mask.

Face crops are extracted per-frame using the RetinaFace pipeline and
resized to $384 \times 384$~pixels prior to model input.

\subsection{Video Clip Construction}
\label{subsec:clips}

Rather than training on individual frames independently, each
training sample is a \emph{video clip}: a temporally ordered sequence
of $T = 4$ frames sampled from a video sequence with stride $s = 2$,
requiring a minimum window of $T \times s = 8$ contiguous frames per
clip. Sequences shorter than this minimum are loop-padded. During
training, the clip start index is sampled uniformly at random from
all valid positions, providing implicit temporal augmentation; at
validation and test time, frame sampling is deterministic.

\subsection{Backbones and Temporal Aggregation}
\label{subsec:architectures}

We evaluate four ImageNet-pretrained CNN backbones: ResNet-50~\cite{he2016resnet},
MobileNetV3-Large via two configurations~(MobileNet from
torchvision~\cite{howard2019mobilenetv3} and MobileNetV3 from
\texttt{timm}~\cite{rw2019timm}), and
EfficientNet-B0~\cite{tan2019efficientnet}. All four share a common
structure: the backbone $f_\theta$ processes each frame independently
to produce a per-frame feature $\mathbf{z}_t \in \mathbb{R}^{B \times d}$,
\begin{equation}
\mathbf{z}_t = f_\theta(\mathbf{x}_{:,t,:,:,:}),
\quad t = 1, \ldots, T,
\label{eq:framefeat}
\end{equation}
which is aggregated across $T$ frames and passed to a linear
classification head. Table~\ref{tab:model_summary} summarises
per-model configurations.

For ResNet-50, MobileNet, and EfficientNet-B0, aggregation is
element-wise mean pooling:
\begin{equation}
\bar{\mathbf{z}} = \frac{1}{T} \sum_{t=1}^{T} \mathbf{z}_t.
\label{eq:meanpool}
\end{equation}
For MobileNetV3, we use a lightweight \emph{Temporal Transformer}:
frame features are projected to $d = 256$, a learnable
\texttt{[CLS]} token is prepended, and the sequence is passed
through a 2-layer transformer encoder with 4~attention heads, GELU
activation, feedforward dimension 512, and dropout 0.1:
\begin{equation}
[\mathbf{v}_\texttt{cls}; \mathbf{v}_1; \ldots; \mathbf{v}_T]
= \text{TransformerEncoder}([\mathbf{e}_\texttt{cls};
\mathbf{z}_1^\prime; \ldots; \mathbf{z}_T^\prime]).
\label{eq:transformer}
\end{equation}
This contrast lets us assess whether learned temporal attention
yields gains over simple averaging at PAD-relevant clip lengths.

\begin{table}[t]
\centering
\renewcommand{\arraystretch}{1.2}
\caption{Architecture summary. $d$ = feature dimension per frame.
\textbf{Mean} = temporal mean pooling;
\textbf{TF} = \texttt{[CLS]}-token Transformer encoder.}
\label{tab:model_summary}
\resizebox{\columnwidth}{!}{%
\begin{tabular}{lcccc}
\hline
\textbf{Model} & \textbf{Pretrained} & \textbf{$d$} &
\textbf{Temporal Agg.} & \textbf{Head} \\
\hline
ResNet-50        & ImageNet (V2) & 2048 & Mean & FC(2048$\to$3) \\
MobileNet        & ImageNet (V1) & 960  & Mean & FC(960$\to$3)  \\
MobileNetV3      & ImageNet      & 256  & TF   & FC(256$\to$3)  \\
EfficientNet-B0  & ImageNet (V1) & 1280 & Mean & FC(1280$\to$3) \\
\hline
\end{tabular}%
}
\end{table}

\subsection{Training Configuration}
\label{subsec:training}

\subsubsection{Data Split and Class Weighting}
The dataset is split at the subject identity level~(80:10:10; 209
train / 26 val / 26 test). Inverse-frequency class weights are
applied to the cross-entropy loss to address class imbalance:
\begin{equation}
w_c = \frac{N_{\text{total}}}{|\mathcal{C}| \cdot N_c},
\quad c \in \{\text{live}, \text{spoof}, \text{props}\}.
\label{eq:classweights}
\end{equation}

\subsubsection{Loss Function}
All models optimise weighted categorical cross-entropy. For
MobileNetV3, label smoothing ($\epsilon = 0.1$) additionally
mitigates overconfidence~\cite{sciopen2025cnn}:
\begin{equation}
\mathcal{L} = -\frac{1}{N}\sum_{i=1}^{N}
\sum_{c \in \mathcal{C}} w_c \,
\tilde{y}_{i,c} \log \hat{p}_{i,c},
\label{eq:loss}
\end{equation}
where $\tilde{y}_{i,c} = (1 - \epsilon)\,y_{i,c}
+ \epsilon / |\mathcal{C}|$ ($\epsilon = 0$ for the remaining
models).

\subsubsection{Optimiser and Schedule}
All models use AdamW~\cite{loshchilov2019adamw} ($\eta =
3\times10^{-4}$, weight decay $10^{-4}$) with cosine annealing:
\begin{equation}
\eta_k = \eta_{\min} + \tfrac{1}{2}(\eta_0 - \eta_{\min})
\!\left(1 + \cos\!\left(\tfrac{k\pi}{K}\right)\right).
\label{eq:cosine}
\end{equation}
MobileNetV3 additionally uses a 5\% linear warmup to stabilise early
gradient flow for the transformer
component~\cite{luevano2024mobilenetv3}.

\subsubsection{Training Details}
ResNet-50, MobileNet, and EfficientNet train for up to 20 epochs
(batch size 4). MobileNetV3 trains for up to 50 epochs with early
stopping~(patience 10) and gradient clipping at norm 1.0. Best
checkpoint selected by minimum validation loss; seed fixed at 42;
mixed-precision~(FP16) used where CUDA is available.

\subsubsection{Data Augmentation}
Training-time augmentations: resize to $384\times384$; random
horizontal flip~(p\,=\,0.5); colour jitter~(brightness,
contrast, saturation~$\pm$0.2; hue~$\pm$0.05, applied with
probability 0.3 for MobileNetV3 and unconditionally for others);
ImageNet normalisation~($\mu = [0.485, 0.456, 0.406]$,
$\sigma = [0.229, 0.224, 0.225]$) for ResNet-50, MobileNet, and
EfficientNet; zero-mean unit-variance for MobileNetV3~(matching
its \texttt{timm} recipe). At validation and test time only
resizing and normalisation are applied.

\subsection{Evaluation Metrics}
\label{subsec:metrics}

Performance is evaluated using ISO/IEC~30107-3
metrics~\cite{iso301073}: \textbf{APCER} (proportion of attacks
classified as bonafide), \textbf{BPCER} (proportion of bonafide
presentations classified as attacks), and their mean \textbf{ACER}:
\begin{equation}
\text{ACER} = \frac{\text{APCER} + \text{BPCER}}{2}.
\label{eq:acer}
\end{equation}
These are supplemented by overall accuracy and per-class
classification rate (live, spoof, props), reported per distance
band, sensor, and environment to isolate failure modes.

%-----------------------------------------------------------------------
\section{Experiments and Results}
\label{sec:experiments}

\subsection{Overall Performance}

Table~\ref{tab:overall_results} reports test-set performance on
22,033 video frames. EfficientNet-B0 achieves the best
accuracy~(98.48\%) and lowest ACER~(2.01\%). MobileNet follows at
97.83\%~(ACER 2.80\%) with near-perfect spoof recall~(99.68\%).
ResNet-50 ties EfficientNet on spoof classification rate~(99.84\%).
MobileNetV3 trails at 91.28\%~(ACER 6.99\%). Across all models,
spoof recall~(95.31--99.84\%) is the highest per-class metric; live
recall is universally lowest, confirming that genuine subjects at
distance are the primary source of false alarms. No model produces
a spoof-to-live error for standalone attacks.

\begin{table}[t]
\centering
\small
\renewcommand{\arraystretch}{1.1}
\caption{Overall test performance~(22,033 frames). Best per column
\textbf{bold}. ACER: lower is better.}
\label{tab:overall_results}
\resizebox{\columnwidth}{!}{%
\begin{tabular}{lccccc}
\hline
\textbf{Model} & \textbf{Acc} & \textbf{R$_\text{live}$}
  & \textbf{R$_\text{spoof}$} & \textbf{R$_\text{props}$}
  & \textbf{ACER} \\
\hline
EfficientNet-B0 & \textbf{98.48} & \textbf{96.51} & 99.84
  & \textbf{99.01} & \textbf{2.01} \\
MobileNet       & 97.83 & 95.28 & 99.68 & 98.52 & 2.80 \\
ResNet-50       & 97.70 & 94.62 & \textbf{99.84} & 98.54 & 2.99 \\
MobileNetV3     & 91.28 & 91.11 & 95.31 & 91.02 & 6.99 \\
\hline
\end{tabular}}
\end{table}

\subsection{Per-Distance and Sensor Analysis}

\begin{figure}[!ht]
  \centering
  \includegraphics[width=\columnwidth]{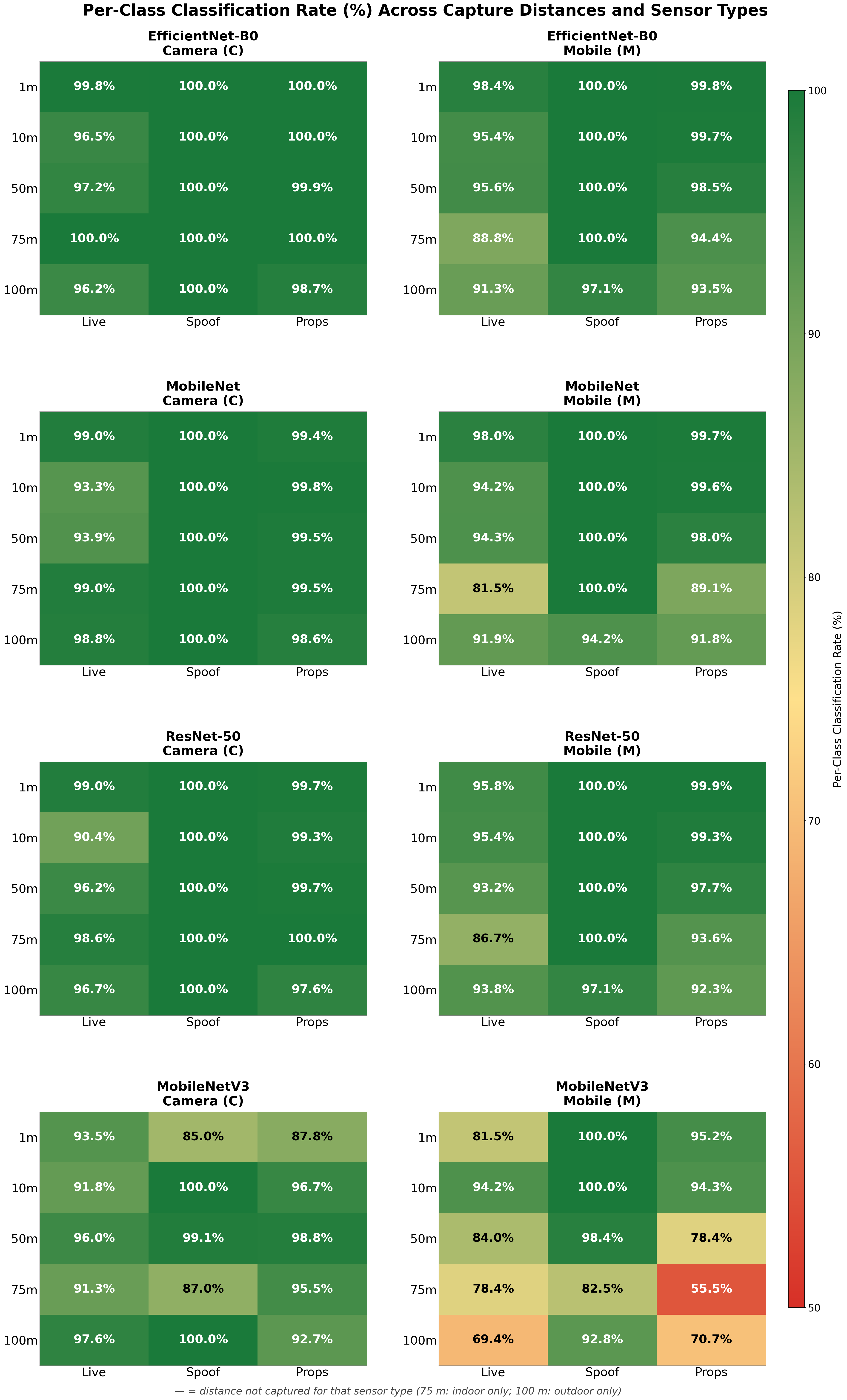}
  \caption{Per-class recall~(\%) across distances and sensor types.
  Camera~(C) panels remain uniformly high. Mobile~(M) panels degrade
  beyond 50\,m~(MobileNetV3: 55.5\% props at 75\,m). Spoof is the
  most distance-robust class. Dashes: 75\,m indoor-only; 100\,m
  outdoor-only.}
  \label{fig:recall_heatmap}
\end{figure}

Table~\ref{tab:distance_results} and Fig.~\ref{fig:recall_heatmap}
show per-distance accuracy by sensor. \textbf{DSLR sensor~(C)}:
EfficientNet achieves 99.9--100.0\% at 1--75\,m. \textbf{Mobile
sensor~(M)}: performance is comparable at close range but degrades
sharply beyond 50\,m. EfficientNet falls to 93.9\% at 75\,m;
MobileNetV3 collapses to 64.9\% at 75\,m~(indoor) and 71.3\% at
100\,m~(outdoor). Spoof recall is the most distance-robust metric:
even MobileNetV3 maintains $>$82.5\% spoof recall at all distances,
while props recall suffers the steepest declines (55.5\% at 75\,m
mobile, MobileNetV3). The sharp mobile degradation beyond 50\,m is
consistent with the low raw pixel counts in
Table~\ref{tab:sensor_resolution}: at 75\,m the mobile sensor
yields an IOD of only 8\,px, falling well below any quality
threshold at which reliable face analysis is expected.

\begin{table}[t]
\centering
\small
\renewcommand{\arraystretch}{1.0}
\caption{Per-distance accuracy~(\%) by sensor.
75\,m: indoor only; 100\,m: outdoor only.
\textbf{Bold}: best per distance--sensor block.}
\label{tab:distance_results}
\resizebox{\columnwidth}{!}{%
\begin{tabular}{llccccc}
\hline
\textbf{Snsr} & \textbf{Model}
  & \textbf{1m} & \textbf{10m} & \textbf{50m}
  & \textbf{75m} & \textbf{100m} \\
\hline
\multirow{4}{*}{C}
 & EfficientNet & \textbf{99.9} & \textbf{99.0} & \textbf{99.3} & \textbf{100.0} & 98.1 \\
 & MobileNet    & 98.8 & 98.2 & 98.2 & 99.8 & \textbf{98.7} \\
 & ResNet-50    & 99.1 & 97.1 & 98.9 & 99.7 & 97.5 \\
 & MobileNetV3  & 96.8 & 95.7 & 98.2 & 94.0 & 94.2 \\
\hline
\multirow{4}{*}{M}
 & EfficientNet & \textbf{99.5} & \textbf{98.7} & \textbf{98.0} & \textbf{93.9} & \textbf{93.2} \\
 & MobileNet    & 99.3 & 98.6 & 97.3 & 91.2 & 92.0 \\
 & ResNet-50    & 98.9 & 98.4 & 96.8 & 92.9 & 92.9 \\
 & MobileNetV3  & 94.5 & 95.4 & 81.0 & 64.9 & 71.3 \\
\hline
\end{tabular}}
\end{table}

\begin{figure}[!ht]
  \centering
  \includegraphics[width=\columnwidth]{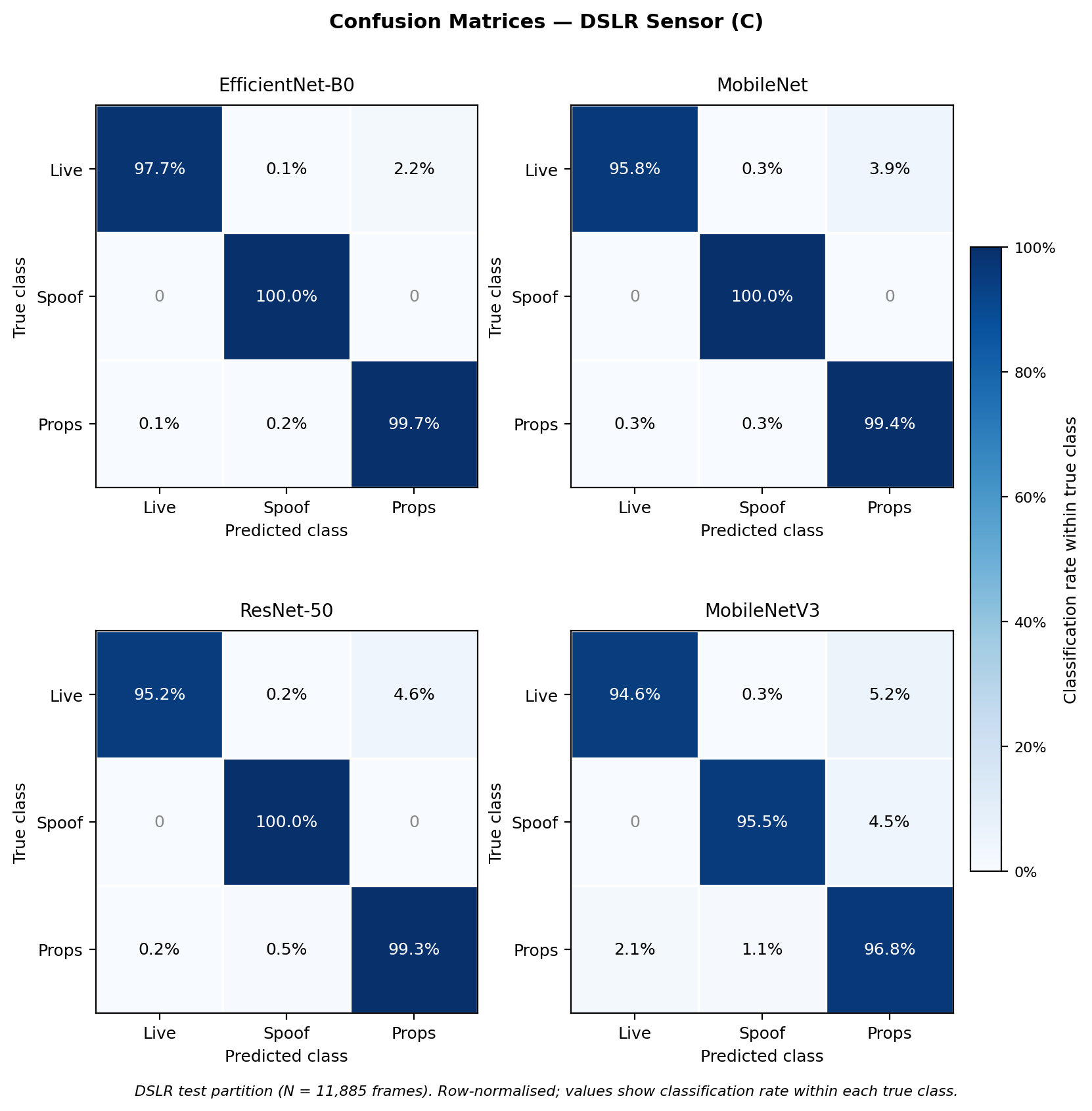}
  \caption{Row-normalised confusion matrices on the DSLR test
  partition~(N\,=\,11{,}885 frames). Spoof$\to$Live is zero for all
  four models.}
  \label{fig:confusion_dslr}
\end{figure}

\begin{figure}[!ht]
  \centering
  \includegraphics[width=\columnwidth]{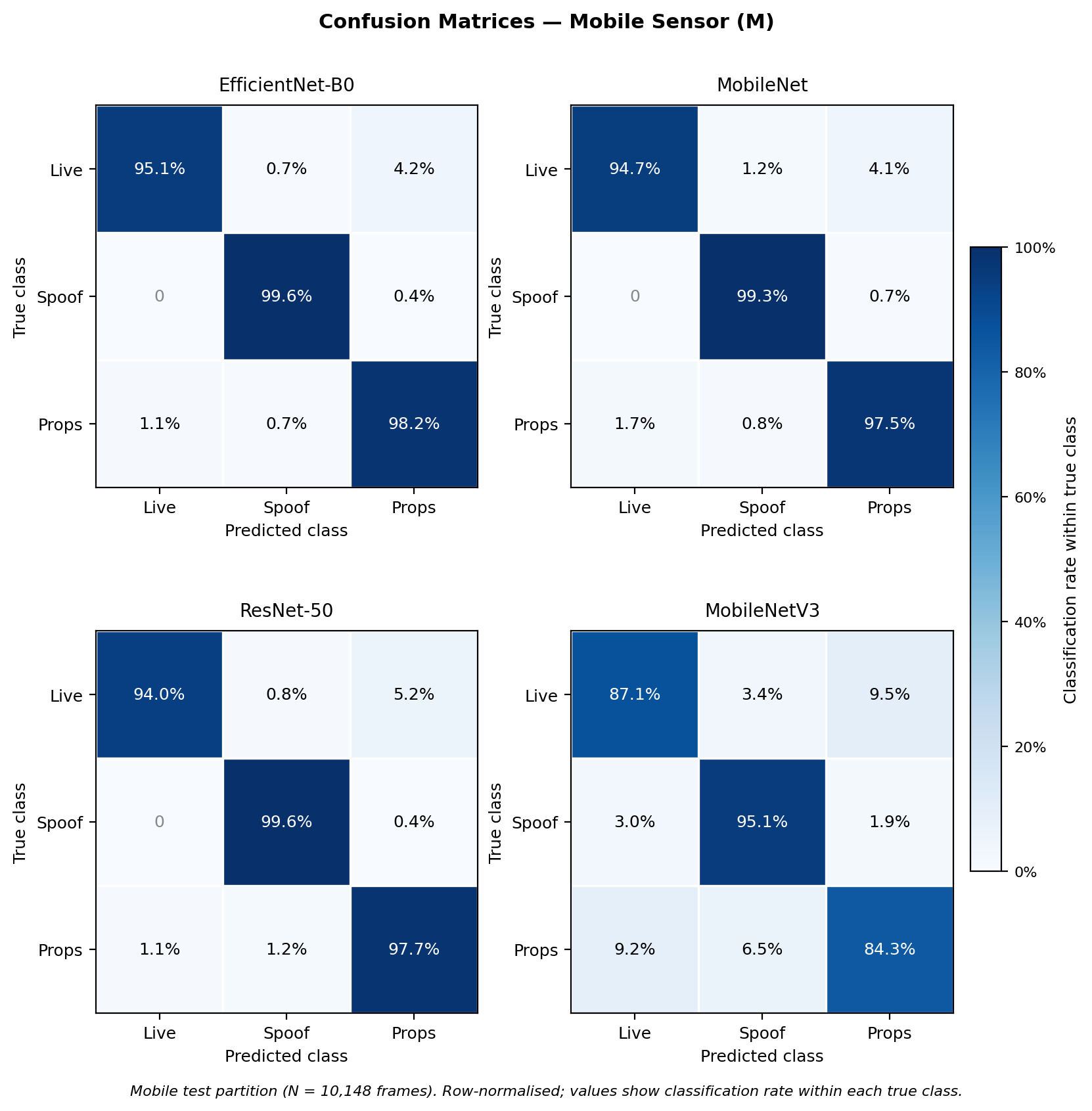}
  \caption{Row-normalised confusion matrices on the mobile test
  partition~(N\,=\,10{,}148 frames). MobileNetV3 shows substantially
  larger off-diagonal mass, particularly for the props class
  (84.3\% vs.\ 96.8\% on the DSLR partition), consistent with the
  low pixel counts at extended distances~(Table~\ref{tab:sensor_resolution}).}
  \label{fig:confusion_mobile}
\end{figure}

\begin{figure*}[!ht]
  \centering
  \includegraphics[width=\textwidth]{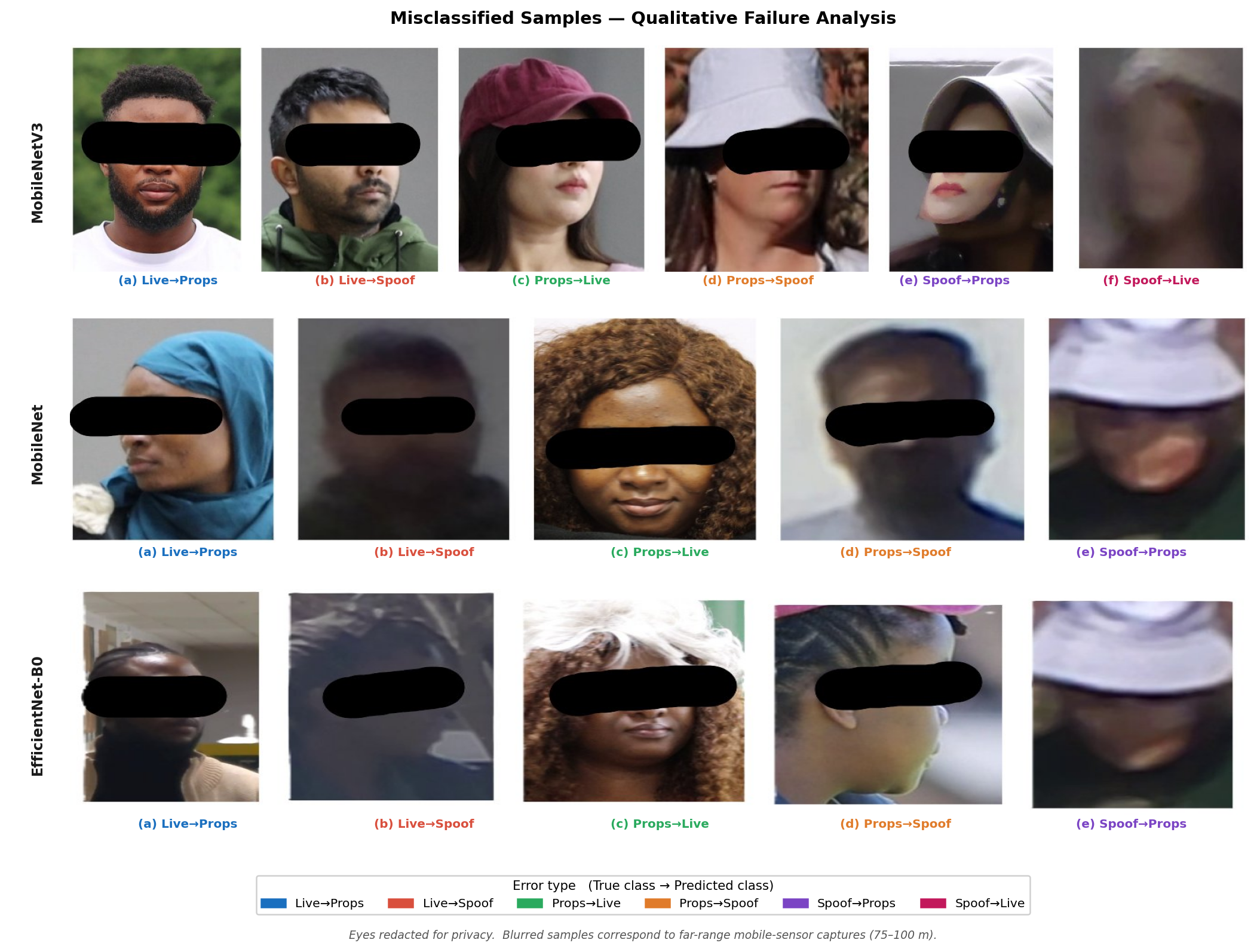}
  \caption{Misclassified samples for MobileNetV3~(top),
  MobileNet~(middle), EfficientNet-B0~(bottom). Eyes redacted.
  Blurred: far-range mobile captures~(75--100\,m). Labels:
  \textit{True\,class\,$\to$\,Predicted\,class}:
  (a)~Live$\to$Props, (b)~Live$\to$Spoof, (c)~Props$\to$Live,
  (d)~Props$\to$Spoof, (e)~Spoof$\to$Props,
  (f)~Spoof$\to$Live~(MobileNetV3 only).}
  \label{fig:misclassifications}
\end{figure*}

\subsection{Environmental and Cross-Dataset Analysis}

Indoor/outdoor accuracy differences are small and non-monotonic,
reflecting the distance-grid asymmetry~(75\,m indoor-only; 100\,m
outdoor-only). EfficientNet: 99.17\% indoor vs.\ 98.01\% outdoor;
MobileNet slightly reverses~(97.42\% vs.\ 98.11\%). Neither
environment is categorically harder; distance and sensor resolution
dominate.

Cross-dataset generalisation tests on three unseen external sets
yield 89--94\% accuracy across all
models~(Table~\ref{tab:generalization}), validating that
distance-diverse training transfers to novel attack categories.

\begin{table}[t]
\centering
\small
\renewcommand{\arraystretch}{1.1}
\caption{Cross-dataset generalisation~(\%).
Set A: chin mask + LivDet + filament~($n$=2,705).
Set B: mouth-chin mask + CelebA + cloth~($n$=2,620).
Set C: complex/simple masks + CelebA + HQ 3D~($n$=4,882).}
\label{tab:generalization}
\begin{tabular}{lccc}
\hline
\textbf{Model} & \textbf{Set A} & \textbf{Set B} & \textbf{Set C} \\
\hline
EfficientNet-B0 & 89 & \textbf{94} & \textbf{93} \\
MobileNet       & 89 & \textbf{94} & \textbf{93} \\
ResNet-50       & \textbf{93} & \textbf{94} & \textbf{93} \\
MobileNetV3     & 89 & 92 & 92 \\
\hline
\end{tabular}
\end{table}

\subsection{Qualitative Failure Analysis}

Figures~\ref{fig:confusion_dslr} and~\ref{fig:confusion_mobile}
show normalised confusion matrices of the predicted classes; while
Fig.~\ref{fig:misclassifications} shows representative failure
cases. Three patterns dominate. (i)~\textbf{Live$\to$Props}: the
most frequent error, driven by far-range mobile captures where
reduced resolution forces reliance on silhouette cues.
(ii)~\textbf{Props$\to$Spoof}: occlusion geometry resembles mask boundaries, operationally the highest-risk confusion.
(iii)~\textbf{Spoof$\to$Props}: compound attacks~(3D mask + props)
occasionally misclassified as props. Critically, no standalone
spoof is ever classified as live by any model.

%-----------------------------------------------------------------------
\section{Discussion}
\label{sec:discussion}

\textbf{Sensor characteristics, not nominal distance alone, govern
PAD failure.} The DSLR sustains $>$94\% accuracy across all five
distances for every model, while the mobile sensor degrades sharply
beyond 50\,m, most severely for MobileNetV3~(64.9\% at 75\,m). The
two sensors differ along several dimensions simultaneously: optical
resolution, sensor area, lens characteristics, and low-light
sensitivity. Our experiments do not isolate which of these factors
dominates; the most parsimonious interpretation is that face-region
pixel count falls below the threshold CNNs require for fine-grained
discrimination, as evidenced by the IOD values in
Table~\ref{tab:sensor_resolution}. This reframes deployment guidance
in sensor terms: any sensor of comparable optical specification to
the DSLR should operate reliably out to 100\,m, while
mobile-grade sensors should restrict PAD to $\leq$50\,m. Controlled
experiments pairing distance with variation in optical resolution,
focal length, and sensor size are needed to establish causality.

\textbf{The props class exposes failure modes invisible to binary
PAD.} Props classification rate degrades faster than spoof at
distance, with MobileNetV3 reaching 55.5\% props recall at
75\,m~(mobile) while maintaining 82.5\% spoof classification rate
at the same distance. In standard binary benchmarks this would be
masked as a single bonafide error rate; the three-class formulation
makes it actionable.

\textbf{Architecture guidance.} EfficientNet-B0 is recommended for
general deployment~(best accuracy, lowest ACER, best live and props
recall). MobileNet offers the strongest efficiency--accuracy
tradeoff for edge deployment. ResNet-50 performs comparably to
MobileNet without an efficiency advantage. MobileNetV3's Temporal
Transformer underperforms mean pooling, likely because 4-frame clips
provide insufficient temporal structure for attention to exploit.

\textbf{Limitations.} This work is a benchmark contribution; the
evaluated backbones are established CNNs chosen to ensure that
observed differences are attributable to dataset factors rather than
architectural novelty. Comparison against transformer-based
architectures~(TimeSformer~\cite{bertasius2021timesformer},
SlowFast~\cite{feichtenhofer2019slowfast}) and generative
anti-spoofing approaches under the same distance-controlled protocol
is an important direction for future work, as is the extension of
clip length and temporal sampling density. The dataset does not
include replay or deepfake attacks; these attack types introduce
additional device-specific confounds at long range and are left to
future dataset expansion. Demographic balance is limited by
university-based recruitment.

%-----------------------------------------------------------------------
\section{Conclusion}
\label{sec:conclusion}

We introduced the first video-based PAD benchmark with five
controlled standoff distances~(1--100\,m), two sensors of differing
optical resolution, indoor/outdoor environments, and a three-class
formulation~(live, spoof, props) that exposes occlusion-specific
failure modes invisible to binary benchmarks. EfficientNet-B0
achieves 98.48\% accuracy and ACER 2.01\%. The DSLR, equipped
with a telephoto zoom lens at 600\,mm sustains near-perfect PAD
at all distances, yielding an IOD of 118\,px at 100\,m that exceeds
NIST\,FRVT quality thresholds; the mobile sensor falls to 64.9\%
accuracy at 75\,m, consistent with its severely reduced pixel count
at extended range. No model misclassifies a standalone spoof as live
at any distance. Cross-dataset generalisation of 89--94\% confirms
that multi-distance, multi-condition training transfers to novel
attack categories. These findings provide direct guidance for
operational PAD deployment in border control and airport
surveillance.

%-----------------------------------------------------------------------
\section*{Ethics Statement}

All 250 participants provided written informed consent prior to data
collection. The study was conducted under institutional ethical
approval IRB. Participants were informed
that their biometric data would be used for research purposes
including scientific publications. Eye regions in all figures are
redacted as an additional privacy measure consistent with the consent
agreement.

\section*{Data Availability}

The dataset will be publicly released to the research community
under an End-User License Agreement~(EULA). The release will include
video clips~(or extracted frame sets), per-frame labels
(live/spoof/props), distance and sensor metadata, and the subject
identity splits used in this paper to enable reproducible evaluation.
Access details are provided at ieee-dataport\cite{pw2z-ya50-26}

\section*{Acknowledgments}

This research was supported by the U.S. Department of
Defense and Ombra. 
This study was conducted under Clarkson University Institutional
Review Board (IRB) approval No. 25-01.3.

%-----------------------------------------------------------------------
{\small
\bibliographystyle{ieee}
\bibliography{egbib}

@article{thapa2023siw,
  author    = {Thapa, Nishan and Chaudhari, Manas and Roy, Kaushik},
  title     = {Presentation Attack Detection: An Analysis of Spoofing in the Wild ({SiW}) Dataset Using Deep Learning Models},
  journal   = {Discover Artificial Intelligence},
  volume    = {3},
  number    = {1},
  pages     = {30},
  year      = {2023},
  doi       = {10.1007/s44163-023-00077-1},
  publisher = {Springer}
}

@inproceedings{wang2023wildface,
  author    = {Wang, Dong and others},
  title     = {Wild Face Anti-Spoofing Challenge 2023: Benchmark and Results},
  booktitle = {Proceedings of the IEEE/CVF Conference on Computer Vision and Pattern Recognition Workshops (CVPRW)},
  pages     = {6380--6391},
  year      = {2023}
}

@inproceedings{zhang2019casia,
  author    = {Zhang, Shifeng and Wang, Xiaobo and Liu, Ajian and Zhao, Chenxu and Wan, Jun and Escalera, Sergio and Shi, Hailin and Wang, Zezheng and Li, Stan Z.},
  title     = {A Dataset and Benchmark for Large-Scale Multi-Modal Face Anti-Spoofing},
  booktitle = {Proceedings of the IEEE/CVF Conference on Computer Vision and Pattern Recognition (CVPR)},
  year      = {2019}
}

@inproceedings{chingovska2012lbp,
  author    = {Chingovska, Ivana and Anjos, Andr{\'e} and Marcel, S{\'e}bastien},
  title     = {On the Effectiveness of Local Binary Patterns in Face Anti-Spoofing},
  booktitle = {2012 BIOSIG -- Proceedings of the International Conference of Biometrics Special Interest Group},
  pages     = {1--7},
  year      = {2012}
}

@inproceedings{boulkenafet2017oulunpu,
  author    = {Boulkenafet, Zinelabidine and Komulainen, Jukka and Li, Lei and Feng, Xiaoyi and Hadid, Abdenour},
  title     = {{OULU-NPU}: A Mobile Face Presentation Attack Database with Real-World Variations},
  booktitle = {2017 12th IEEE International Conference on Automatic Face \& Gesture Recognition},
  year      = {2017}
}

@article{sciopen2025cnn,
  author    = {Al-Smadi, Mohammad and others},
  title     = {Optimizing {CNN} Architectures for Face Liveness Detection: Performance, Efficiency, and Generalization across Datasets},
  journal   = {Computer Modeling in Engineering \& Sciences},
  volume    = {143},
  number    = {3},
  year      = {2025},
  doi       = {10.32604/cmes.2025.058855},
  publisher = {Tech Science Press}
}

@inproceedings{luevano2024mobilenetv3,
  author    = {Luevano, Luis and others},
  title     = {Assessing the Performance of Efficient Face Anti-Spoofing Detection Against Physical and Digital Presentation Attacks},
  booktitle = {Proceedings of the IEEE/CVF Conference on Computer Vision and Pattern Recognition Workshops (CVPRW): Face Anti-Spoofing Workshop},
  year      = {2024}
}

@article{etasr2025flnet,
  author    = {Ibrahim, Samar M. and Ibrahim, Muhammad S. and Khan, Salman and Ko, Young Woong and Lee, Jong Gook},
  title     = {Novel Deep {CNN} Architectures for Anti-Spoofing},
  journal   = {Engineering, Technology \& Applied Science Research},
  volume    = {15},
  number    = {5},
  pages     = {27206--27212},
  year      = {2025},
  doi       = {10.48084/etasr.8229}
}

@inproceedings{lucena2017transfer,
  author    = {Lucena, Oeslle and Junior, Amadeu and Moia, Vitor and Souza, Roberto and Valle, Eduardo and Lotufo, Roberto},
  title     = {Transfer Learning Using Convolutional Neural Networks for Face Anti-Spoofing},
  booktitle = {Image Analysis and Recognition: 14th International Conference, ICIAR 2017},
  pages     = {27--34},
  year      = {2017},
  publisher = {Springer}
}

@inproceedings{george2020multichannel,
  author    = {George, Anjith and Mostaani, Zohreh and Geissenbuhler, David and Nikisins, Olegs and Anjos, Andr{\'e} and Marcel, S{\'e}bastien},
  title     = {Biometric Face Presentation Attack Detection With Multi-Channel Convolutional Neural Network},
  journal   = {IEEE Transactions on Information Forensics and Security},
  volume    = {15},
  pages     = {42--55},
  year      = {2020},
  doi       = {10.1109/TIFS.2019.2916652}
}

@inproceedings{atoum2017patch,
  author    = {Atoum, Yousef and Liu, Yaojie and Jourabloo, Amin and Liu, Xiaoming},
  title     = {Face Anti-Spoofing Using Patch and Depth-Based {CNN}s},
  booktitle = {Proceedings of the IEEE International Joint Conference on Biometrics (IJCB)},
  address   = {Denver, CO},
  year      = {2017}
}

@inproceedings{liu2023unsupervised,
  author    = {Liu, Yuchen and Chen, Yuhao and Gou, Mingming and Huang, Chia-Ting and Wang, Yanmin and Dai, Wei and Xiong, Hongkai},
  title     = {Towards Unsupervised Domain Generalization for Face Anti-Spoofing},
  booktitle = {Proceedings of the IEEE/CVF International Conference on Computer Vision (ICCV)},
  year      = {2023}
}

@inproceedings{zhou2023iadg,
  author    = {Zhou, Qianyu and Zhang, Ke-Yue and Yao, Taiping and Lu, Xuequan and Yi, Ran and Ding, Shouhong and Ma, Lizhuang},
  title     = {Instance-Aware Domain Generalization for Face Anti-Spoofing},
  booktitle = {Proceedings of the IEEE/CVF Conference on Computer Vision and Pattern Recognition (CVPR)},
  pages     = {20453--20463},
  year      = {2023}
}

@inproceedings{jia2020ssdg,
  author    = {Jia, Yunpei and Zhang, Jie and Shan, Shiguang and Chen, Xilin},
  title     = {Single-Side Domain Generalization for Face Anti-Spoofing},
  booktitle = {Proceedings of the IEEE/CVF Conference on Computer Vision and Pattern Recognition (CVPR)},
  pages     = {8484--8493},
  year      = {2020}
}

@inproceedings{cornett2023briar,
  author    = {Cornett, David and Brogan, Joel and Barber, Nell and Aykac, Deniz and Baird, Seth and Burchfield, Nicholas and Dukes, Carl and Duncan, Abby and Ferrell, Regina and Goddard, Joel and others},
  title     = {Expanding Accurate Person Recognition to New Altitudes and Ranges: The {BRIAR} Dataset},
  booktitle = {Proceedings of the IEEE/CVF Winter Conference on Applications of Computer Vision Workshops (WACVW)},
  pages     = {593--602},
  year      = {2023}
}

@article{farsight2025,
  author    = {Santos-Villalobos, Hector J. and others},
  title     = {Person Recognition at Altitude and Range: Fusion of Face, Body Shape and Gait},
  journal   = {arXiv preprint arXiv:2505.04616},
  year      = {2025}
}

@article{survfas2023,
  author    = {Fang, Han and Liu, Ajian and Yuan, Haocheng and Zheng, Junqi and Zeng, Dingjie and others},
  title     = {Surveillance Face Anti-Spoofing},
  journal   = {arXiv preprint arXiv:2301.00975},
  year      = {2023}
}

@inproceedings{cvpr2023fascvpr,
  author    = {Face Anti-Spoofing Workshop Organizers},
  title     = {Face Anti-Spoofing Workshop and Challenge at {CVPR2023}},
  booktitle = {Proceedings of the IEEE/CVF Conference on Computer Vision and Pattern Recognition Workshops (CVPRW)},
  year      = {2023},
  note      = {\url{https://sites.google.com/view/face-anti-spoofing-challenge}}
}

@article{regulaforensics2025biometrics,
  author    = {{Regula Forensics}},
  title     = {How Biometrics Is Revolutionizing the Airport Security and Boarding Process},
  journal   = {Regula Forensics Blog},
  year      = {2025},
  note      = {Citing 2024 {IATA} Global Passenger Survey data.
               Available at \url{https://regulaforensics.com/blog/biometrics-in-airport/}}
}

@techreport{dhsoig2024,
  author      = {{U.S. Department of Homeland Security, Office of Inspector General}},
  title       = {{CBP} Does Not Have the Technology to Collect Biometrics from
                 Travelers Arriving in Vehicles at Land Ports of Entry},
  institution = {DHS Office of Inspector General},
  year        = {2024},
  note        = {OIG Report. Available at \url{https://www.oig.dhs.gov}}
}

@article{wen2015nuaa,
  author    = {Wen, Di and Han, Hu and Jain, Anil K.},
  title     = {Face Spoof Detection With Image Distortion Analysis},
  journal   = {IEEE Transactions on Information Forensics and Security},
  volume    = {10},
  number    = {4},
  pages     = {746--761},
  year      = {2015},
  doi       = {10.1109/TIFS.2015.2400395}
}

@inproceedings{he2016resnet,
  author    = {He, Kaiming and Zhang, Xiangyu and Ren, Shaoqing and Sun, Jian},
  title     = {Deep Residual Learning for Image Recognition},
  booktitle = {Proceedings of the IEEE Conference on Computer Vision and Pattern Recognition (CVPR)},
  pages     = {770--778},
  year      = {2016}
}

@inproceedings{howard2019mobilenetv3,
  author    = {Howard, Andrew and Sandler, Mark and Chu, Grace and Chen, Liang-Chieh and Chen, Bo and Tan, Mingxing and Wang, Weijun and Zhu, Yukun and Pang, Ruoming and Vasudevan, Vijay and Le, Quoc V. and Adam, Hartwig},
  title     = {Searching for {MobileNetV3}},
  booktitle = {Proceedings of the IEEE/CVF International Conference on Computer Vision (ICCV)},
  pages     = {1314--1324},
  year      = {2019}
}

@inproceedings{tan2019efficientnet,
  author    = {Tan, Mingxing and Le, Quoc V.},
  title     = {{EfficientNet}: Rethinking Model Scaling for Convolutional Neural Networks},
  booktitle = {Proceedings of the International Conference on Machine Learning (ICML)},
  pages     = {6105--6114},
  year      = {2019}
}

@techreport{iso301073,
  author      = {{ISO/IEC}},
  title       = {Information Technology -- Biometric Presentation Attack Detection -- Part 3: Testing and Reporting},
  institution = {International Organization for Standardization},
  number      = {ISO/IEC 30107-3},
  year        = {2023}
}

@article{wang2025fassurvey,
  author    = {Wang, Hao and others},
  title     = {Face Anti-Spoofing Based on Deep Learning: A Comprehensive Survey},
  journal   = {Applied Sciences},
  volume    = {15},
  number    = {12},
  pages     = {6891},
  year      = {2025},
  doi       = {10.3390/app15126891},
  publisher = {MDPI}
}

@article{retinaface2019,
  author    = {Deng, Jiankang and Guo, Jia and Zhou, Yuxiang and Yu, Jinke
               and Kotsia, Irene and Zafeiriou, Stefanos},
  title     = {{RetinaFace}: Single-Stage Dense Face Localisation in the Wild},
  journal   = {arXiv preprint arXiv:1905.00641},
  year      = {2019}
}

@misc{21362714,
  author = {Chowdhury, Mohammad Zahir Uddin},
  title  = {Face Detection by {RetinaFace}},
  year   = {2026},
  month  = jul,
  note   = {\url{https://doi.org/10.5281/zenodo.21362714}}
}

@techreport{nistfrvt2020,
  title = {Face Recognition Vendor Test (FRVT) Part 3: Demographic Effects},
  author = {NIST},
  year = {2020},
  institution = {National Institute of Standards and Technology},
  number = {NISTIR 8280},
  url = {https://doi.org/10.6028/NIST.IR.8280}
}

@software{opencv2015,
  author = {{OpenCV team}},
  title  = {{OpenCV}: Open Source Computer Vision Library},
  url    = {https://opencv.org},
  year   = {2015}
}

@misc{rw2019timm,
  author    = {Wightman, Ross},
  title     = {{PyTorch Image Models (timm)}},
  year      = {2019},
  howpublished = {\url{https://github.com/huggingface/pytorch-image-models}},
  note      = {GitHub repository}
}

@inproceedings{loshchilov2019adamw,
  author    = {Loshchilov, Ilya and Hutter, Frank},
  title     = {Decoupled Weight Decay Regularization},
  booktitle = {Proceedings of the International Conference on
               Learning Representations (ICLR)},
  year      = {2019}
}

@inproceedings{bertasius2021timesformer,
  author    = {Bertasius, Gedas and Wang, Heng and Torresani, Lorenzo},
  title     = {Is Space-Time Attention All You Need for Video Understanding?},
  booktitle = {Proceedings of the International Conference on
               Machine Learning (ICML)},
  pages     = {813--824},
  year      = {2021}
}

@inproceedings{feichtenhofer2019slowfast,
  author    = {Feichtenhofer, Christoph and Fan, Haoqi and Malik, Jitendra
               and He, Kaiming},
  title     = {{SlowFast} Networks for Video Recognition},
  booktitle = {Proceedings of the IEEE/CVF International Conference on
               Computer Vision (ICCV)},
  pages     = {6202--6211},
  year      = {2019}
}

@data{pw2z-ya50-26,
doi = {10.21227/pw2z-ya50},
url = {https://dx.doi.org/10.21227/pw2z-ya50},
author = {Mohammad Zahir Uddin Chowdhury and Lambert Igene and Avery Shoemaker and Kolade Ayeni and Aniekan-Abasi Inyang and Paige Estabrook and Parker Isaacson and Valence Yagayandi and Jonathan Welbeck and Peter Johnson and Stephanie Schuckers},
publisher = {IEEE Dataport},
title = {MORSE-Face Dataset: A Multi-Distance, Occlusion-Rich, Real and Spoof Evaluation Face Dataset},
year = {2026} }

@inproceedings{igene2024development,
  title={Development of novel 3d face masks for assessment of face recognition systems},
  author={Igene, Lambert and Hossain, Afzal and Plesh, Richard and Purnapatra, Sandip and Singh, Surendra and Imtiaz, Masudul and Schuckers, Stephanie},
  booktitle={2024 International Conference of the Biometrics Special Interest Group (BIOSIG)},
  pages={1--7},
  year={2024},
  organization={IEEE}
}

@inproceedings{igene2024face,
  title={Face liveness detection competition (livdet-face)-2024},
  author={Igene, Lambert and Hossain, Afzal and Chowdhury, Mohammad Zahir Uddin and Rezaie, Humaira and Rollins, Ayden and Dykes, Jesse and Vijaykumar, Rahul and Komaty, Alain and Marcel, S{\'e}bastien and Schuckers, Stephanie and others},
  booktitle={2024 IEEE International Joint Conference on Biometrics (IJCB)},
  pages={1--9},
  year={2024},
  organization={IEEE}
}
}

\end{document}